# Improvement of human health lifespan with hybrid group pose estimation methods

Arindam Chaudhuri[1]

[1] Samsung R & D Institute Noida 201304 Delhi India
`arindam_chau@yahoo.co.in; arindamphdthesis@gmail.com`

**Abstract.** Human beings rely heavily on estimation of poses in order to access their body movements. Human pose estimation methods take advantage of computer vision advances in order to track human body movements in real life applications. This comes from videos which are recorded through available devices. These paradigms provide potential to make human movement measurement more accessible to users. The consumers of pose estimation movements believe that human poses content tend to supplement available videos. This has increased pose estimation software usage to estimate human poses. In order to address this problem, we develop hybrid-ensemble-based group pose estimation method to improve human health. This proposed hybrid-ensemble-based group pose estimation method aims to detect multi-person poses using modified group pose estimation and modified real time pose estimation. This ensemble allows fusion of performance of stated methods in real time. The input poses from images are fed into individual methods. The pose transformation method helps to identify relevant features for ensemble to perform training effectively. After this, customized pre-trained hybrid ensemble is trained on public benchmarked datasets which is being evaluated through test datasets. The effectiveness and viability of proposed method is established based on comparative analysis of group pose estimation methods and experiments conducted on benchmarked datasets. It provides best optimized results in real-time pose estimation. It makes pose estimation method more robust to occlusion and improves dense regression accuracy. These results have affirmed potential application of this method in several real-time situations with improvement in human health life span.

## 1   Introduction

In current digital media age, huge data volumes are produced on social media platforms on daily basis. This media data is continuously created, viewed, modified and distributed through electronic devices which doubles almost every month. These massive data chunks are used by researchers regularly for analysis and decision making. The majority of this data comprises of images and videos. This visual content is always more appreciable and memorable for people [1]. However, there remains a challenge in understanding available objects in these data. This process of understanding objects in images and videos is known as pose estimation. It is an important activity in artificial intelligence and computer vision. It involves tracking position, detection and human body parts orientation in images and videos [2]. The accurate detection and tracking of human movements call for their quantitative measurements [3]. Some common examples include scrutiny by sports' judges and performance of figure skaters', measures by physical therapist to access patient's speed, inspections by running coaches etc. These movements are interpreted by humans in order to communicate and make emotional state inferences through body language reading [4].

Human pose estimation data can have either single person or multiple persons. The multiple persons pose estimation has been a major source of research attention in information processing for decades. There has been wide spectrum of applications [5], [6] in areas of augmented reality, human computer interaction and virtual reality. The most popular pose estimation applications [7], [8] in past few years include human activity estimation, motion transfer, motion capture for training robots and motion tracking for consoles. Considering an image, prime objective is to localize 2D keypoint positions for every person in image. Several methods have been developed in this area [9], [10], [11], [12], [13]. Inspite of this it remains challenging and intractable problem for situations with heavy occlusions, hard poses and diverse body part scales. Human pose estimation has often been considered as method which helps in measurement of human movement kinematics. Pose estimation methods helps people to identify important landmarks in human body. These are recorded through devices as shown in Fig. 1. The pose estimation task needs to focus at local as well as global dependencies. The local and global dependencies work for human and keypoint levels respectively. These dependencies lead to lead to normalization of body parts. Here concentration is attributed to semantic granularity. The solutions are basically two stage methods which divide problem into two separate sub-problems such as global person detection and local keypoint regression.

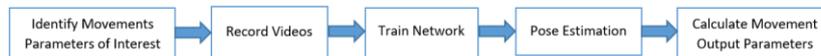

**Fig. 1** The movement kinematics measured with basic pose estimation workflow

Considering the success received from recent end-to-end object detection methods [16], there is a surge of related methods which regard human pose estimation as direct set prediction problem. The two stage methods or DETR [14] suffers because of slow convergence in training. In order to achieve high precision more epochs are needed. In [15] a simple and effective transformer based method for end-to-end multi-person pose estimation is presented. In [16] off-the-shelf detector obtains bounding boxes where individual estimation of pose is performed for each person. In real world pose estimation methods can be used for purpose of entertainment [17]. Pose estimation can also be used to introduce better innovative ideas in movies. It can also be used for malignant purposes which influences socio cultural outcomes.

This research concentrates to develop hybrid ensemble based group pose estimation method to identify human poses with transformation involved. Here pose detection method is considered as regression problem. Based on motivation from [18] and [19], a novel hybrid ensemble based group pose (HEGPosEs) [20] for end-to-end multi-person pose estimation is presented. This computational system consists of (a) modified group pose estimation (MGPosEs) and (b) modified real-time multi-person pose estimation (MRTMPPosEs). MGPosEs and MRTMPPosEs focus on estimating poses at individual levels. The outputs received from these methods are subjected to pose transformation which helps to identify relevant ensemble features such that training is effectively performed. Then pre-trained customized hybrid ensemble is trained on public benchmarked datasets. With respect to various parameters ensemble performance is assessed on test datasets. HEGPosEs is an end-to-end real time method. In order to validate strength of this method two benchmarked datasets are used viz DensePose-COCO [21] and MPII Human Pose [10]. In this research feasibility is provided with respect to occlusion which improves dense regression accuracy. Alongwith this certain optimization based benefits are also provided. This results in more profitable, efficient and sustainable multi-person pose estimation based on several benchmarks with improvement in human health life span. We have structured this paper as follows. In section 2, work done in pose estimation is presented comprising of state-of-the-art group pose estimation methods. The different components of proposed methodology are highlighted in section 3. In section 4, proposed method is illustrated with respect to contextual dataset. Then experiments and analysis are discussed in section 5. Finally, in section 6 conclusion is given.

## 2    Related Work

There has been a constant increase in readily available softwares which have helped in pose estimation. This has raised questions on credibility of poses obtained [22]. The situation becomes more challenging because of human visual appearance, variability in images, lighting conditions, human physique, self-articulation based partial occlusions, object layering in scene, complex human skeletal structures etc. Pose estimation automatically tracks anatomical landmarks also known as keypoints of human body from videos. It is categorized as top-down and bottom-up methods. Top-down methods have body detectors which help in determination of body joints with bounding boxes. Bottom-up methods perform evaluation of each body joint and compose poses which is having unique nature. The primary output of any pose estimation consists of two-dimensional pixel coordinates in series from detected keypoints. There are different methods from 2D pixel coordinates [18], [19], [20], [23]. The 3D human movement kinematics are reconstructed from videos with many viewpoints [24]. During past decade, human pose estimation in multi-person situations has attracted high interest among researchers [4], [5], [6], [20]. The two stage non-end-to-end methods include top down [21], [16], [25] and bottom up methods [26], [27]. The current end-to-end multi-person pose estimation methods are developed considering DETR [14] designs and its variants [28], [29], [30], [31]. PETR [13] views activity as hierarchical set prediction problem. QueryPose [32] and EDPose [33] are adapted to end-to-end method which takes support from Sparse R-CNN. Group Pose [18] adopts simple transformer decoder with improved performance. Earlier pose estimation methods consider keypoint localization as coordinate regression [34], [35] or heatmap regression [9], [31]. Transformer based architectures [36] have been successful in various vision applications such as semantic segmentation, video understanding and pose estimation.

## 3   Designing Group Pose Estimation

Here we give description of proposed pose estimation method. Fig. 2 shows detailed HEGPosEs architecture. In order to detect robust poses, HEGPosEs [20] analyses different human body movements. It is an ensemble of MGPosEs and MRTMPPosEs. Both of these models are optimized for variety of poses. These are capable of modeling [4] variety of human body configurations. HEGPosEs has wide variety of applications in multi-person human pose estimation [20] such as activity recognition, motion capture etc. It does not estimate poses accurately with direct inputs. It is not able to identify features well in isolation. In order to achieve better results from HEGPosEs, proper feature extraction is required. This is achieved with pose transformation method which is performed with PoseTrans [37].

Fig. 3 shows PoseTrans from where new training samples are generated with diverse posesets. PoseTrans consists of transformation alongwith discriminator $PD$ and clustering (PCM) modules. The implausible samples are filtered out with discriminator module to maintain plausibility. PoseTrans uses transformation module till plausible poses are developed. Clustering module clusters poses into several categories. The best pose is selected to be added as new training sample. With clustering poses in datasets, many clusters contain fewer examples. The transformation and discriminator modules address these problems.

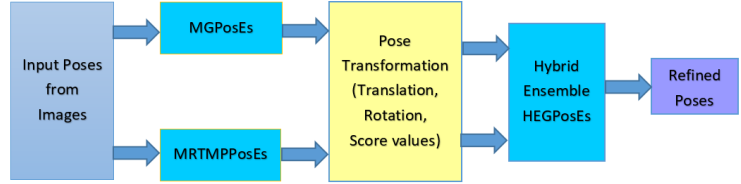

**Fig. 2** HEGPosEs architecture to identify most efficient pose transformation method

The outputs from MGPosEs and MRTMPPosEs are transformed and passed to HEGPosEs. The various steps are shown in Fig. 4. Here objective is to produce best transformation results. We divide transformed pose images into ratio of 80:20 to perform training and validation. With 50 epochs and batch size of 200 this model is trained and validated. The evaluation metrics from confusion matrix are used to validate performance of model.

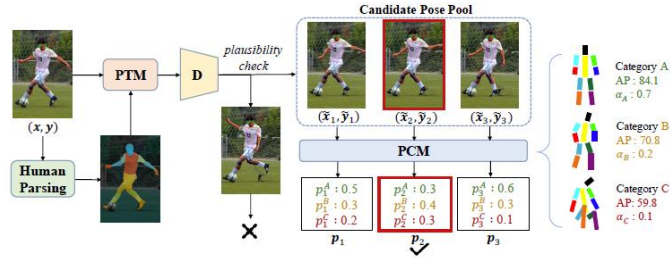

**Fig. 3** PoseTrans generates new training samples with diverse posesets

Benchmarked datasets [10], [21], [38] are used to increase model performance. The model's performance is further improved with transfer learning [20]. The transfer learning helps to identify weightage factor to each individual pose estimation components. In HEGPosEs, pre-trained models are used simultaneously for transfer learning. The domain specific classification backbone used here is ResNet50 [39] and task specific keypoint model is Mask R-CNN [40]. The average performance on target domain is achieved through task-specific model. The feature concatenation is performed through shallow feature converters learning. Another challenge is reaching acceptable training time. It is seen that sometimes training results in models overfit. This is resolved through early stopping. In Fig. 5 steps are demonstrated which incorporates transfer learning with HEGPosEs.

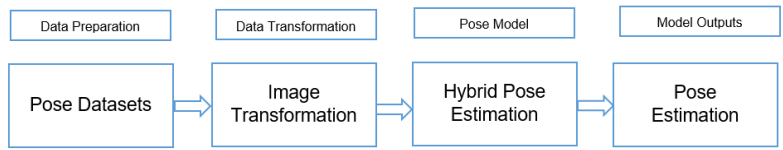

**Fig. 4** Block diagram for pose transformation method identification

Identification of best deep learning method for pose estimation is difficult. Keeping this in view, method proposed learns when stated algorithm is superior with respect to another algorithm. This helps to develop robust multi-person ensemble method. This ensemble allows fusion performance of different methods. The ensemble is created with bagging and stacking methods. Both ensembles depend on set of deep learning models. In each instance output of different methods are considered. The results are further tuned through refinement of obtained human poses. Several deep architectures are trained to develop reliable results with customized vanilla pose estimation ensembles. The ensemble strategies help to develop refined poses. Each ensemble is optimized for variety of purposes. A large number of human body configurations are modelled here. Here prime concentration is provided on accuracy of pose estimation prediction pipeline. Table 1 summarizes hyperparameter values in this method.

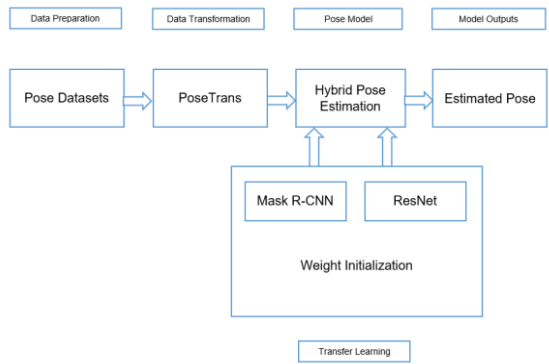

**Fig. 5** Proposed method for hybrid pose estimation

The method is benchmarked with respect to certain baseline models [18], [19]. This helps in identification of performance gain or loss. The validation is performed through metrics multi-task loss and reconstruction loss [20]. The bagging and stacking methods also helps us to arrive at optimum results of refined pose. The bagging methods are defined through mathematical integration of individual results of MGPosEs and MRTMPosEs. These models are considered as base level models. The stacking methods reply on certain statistical and machine learning methods to refine pose results obtained from bagging. They increase predictive performance of ensemble through consideration of best results achieved from bagging. This reduces bias and variance, increases model variety, and improves interpretability factor of final prediction.

| Hyperparameters | Values |
|---|---|
| Dropout rate | 0.4 |
| Learning rate | 0.004 |
| Decay | 0 |
| Epochs | 40 |
| Batch size | 200 |

**Table 1** Hyperparameter values for HEGPosEs

Here bagging ensemble is represented as simple bagging and weighted bagging which take output poses of MGPosEs and MRTMPPosEs to calculate refined pose. The simple bagging method treats each model equally and produces mean of input poses. The resulting translation and rotation are obtained with:

$$Ta_r = \frac{1}{n}\sum_{i=1}^{n} Ta_i \quad (1)$$
$$Ro_r = \text{argmin}_{R \in SOR(3)} \sum_{i=1}^{n} \|R_i - R\|^2 \quad (2)$$

Here chordal L2 mean is used which minimizes square of difference between rotation matrices. In (1) $Ta_r$ represents refined translation and in (2) $Ro_r$ represents refined rotation. The weighted bagging addresses limitations of simple bagging. It considers scores which each model contributes towards its estimation. These scores describe confidence levels of models and detection class. The poses are weighted considering scores with respect to total scores. Each model's weight calculation is done with:

$$wh_i = \frac{1}{(1-sc)^2 + \varepsilon} \quad (3)$$

In (3) $sc$ represents score value achieved and $\varepsilon$ is small positive number to avoid division by zero. The refined translation and rotation are obtained with:

$$Ta_r = \frac{1}{n}\sum_{i=1}^{n} wh_i \cdot Ta_i \quad (4)$$
$$Ro_r = \text{argmin}_{R \in SOR(3)} \sum_{i=1}^{n} wh_i \|R_i - R\|^2 \quad (5)$$

The stacking ensemble is a generalization method which integrates different models [22]. The training of base models is performed followed by validation. It generates new dataset which are independent from each model's output. With validation data integration model is trained. Fig. 6 shows training and evaluation pipeline of model. Integration level methods integrate output of base level model results. Here integration level methods used are Ridge Linear Regression or

L2 Regularization, Random Forests, XGBoost, Regularized Support Vector Regression and Regularized Multi-Layer Perceptron.

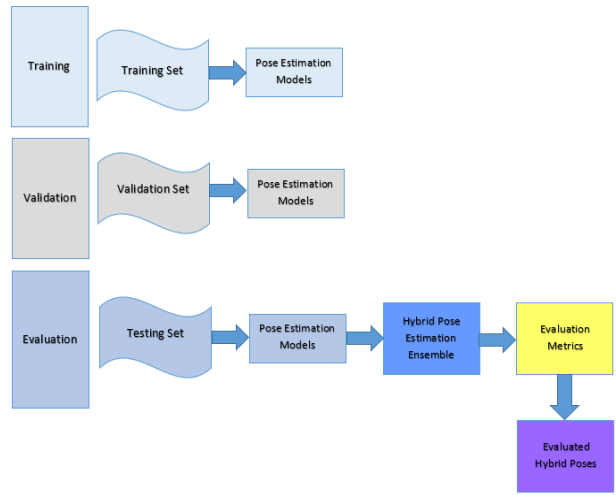

**Fig. 6** Hybrid ensemble model - training, validation and evaluation pipeline

## 4  Illustration of Group Pose Estimation Methodology

Now we illustrate proposed group pose estimation method with DensePose-COCO [21] dataset. It is prepared by manually annotating 500 COCO based images. DensePose-COCO dataset has variety of annotations in alignment with dense correspondences happening between 2D images to surface representations. Annotations and manual labeling are done here with SMPL model and SURREAL textures. This dataset has thousands of multi-person images. It uses ideas from bounding boxes, object detection, image segmentation and other methods. This method is tested with baselines which have semantically meaningful pose estimates. The domain experts provide their viewpoints from poses. The process flow is highlighted in Fig. 7 with example from DensePose-COCO dataset [21].

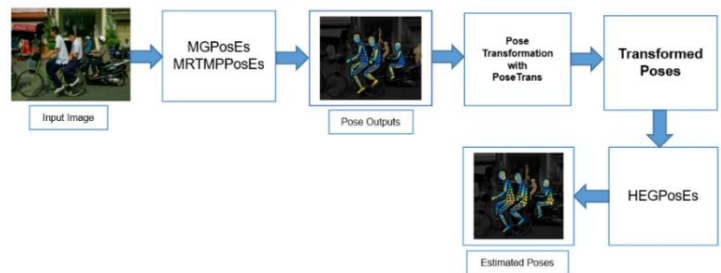

**Fig. 7** Hybrid pose estimation method with DensePose-COCO dataset

A close examination of PoseTrans counterparts for actual and pose images shows that there is uniformity in poses estimated for both image categories. Any difference in estimated poses helps us to differentiate authentic poses from inaccurate poses. The customized HEGPosEs takes transformed images as inputs. With 50 epochs and batch size of 20 we train and validate this model. In Table 2 we summarise model performance with respect to training accuracy, training loss, validation loss, validation accuracy, precision, recall and F1-score. In Fig. 8 performance matrix is presented.

| Training Accuracy | Training Loss | Validation Loss | Validation Accuracy | Precision | Recall | F1 Score |
|---|---|---|---|---|---|---|
| 0.9994 | 0.02 | 0.02 | 0.9992 | 0.9950 | 1.0 | 0.9975 |

**Table 2** The model performance of HEGPosEs on DensePose-COCO dataset

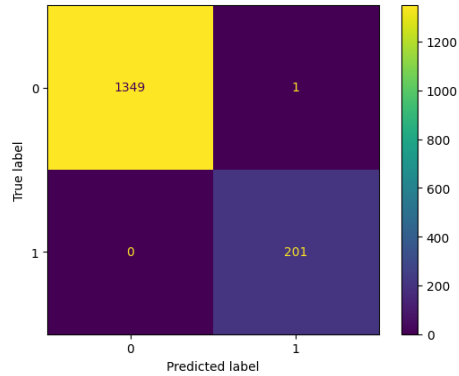

**Fig. 8** Performance matrix for DensePose-COCO datasets

| Training Accuracy | Training Loss | Validation Loss | Validation Accuracy | Precision | Recall | F1 Score |
|---|---|---|---|---|---|---|
| 0.9967 | 0.02 | 0.02 | 0.9964 | 0.9950 | 0.9800 | 0.9874 |

**Table 3** The model performance of HEGPosEs on MPII Human Pose datasets

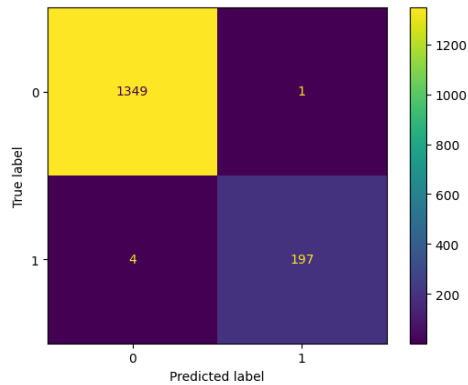

**Fig. 9** Performance matrix for MPII Human Pose datasets

| Method | Backbone | Loss | mAP | mAP_50 | mAP_75 | mAP_Medium | mAP_Large |
|---|---|---|---|---|---|---|---|
| PETR [13] | ResNet-50 | HM + KR | 67.7 | 87.6 | 76.4 | 62.9 | 77.9 |
| PETR [13] | Swin-L | HM + KR | 73.4 | 90.9 | 81.0 | 66.6 | 80.8 |
| QueryPose [32] | ResNet-50 | BR + RLE | 68.9 | 88.7 | 74.5 | 63.9 | 76.6 |
| ED-Pose [33] | ResNet-50 | BR + KR | 71.7 | 89.7 | 78.6 | 66.0 | 79.9 |
| ED-Pose [33] | Swin-L | BR + KR | 75.5 | 94.0 | 81.9 | 68.9 | 82.9 |
| GroupPose [18] | ResNet-50 | KR | 73.0 | 89.5 | 79.6 | 66.9 | 79.9 |
| GroupPose [18] | Swin-T | KR | 73.9 | 90.5 | 80.6 | 68.8 | 81.6 |
| GroupPose [18] | Swin-L | KR | 75.9 | 90.7 | 82.7 | 69.6 | 84.0 |
| RTMPose-t [19] | CSPNeXt-t | KR | 65.5 | 88.6 | 75.6 | 68.8 | 77.0 |
| RTMPose-s [19] | CSPNeXt-s | KR | 68.9 | 90.9 | 78.9 | 67.0 | 80.0 |
| RTMPose-m [19] | CSPNeXt-m | KR | 73.6 | 88.9 | 80.8 | 69.9 | 84.8 |
| RTMPose-l [19] | CSPNeXt-l | KR | 75.6 | 90.0 | 80.9 | 67.9 | 80.9 |
| RTMPose-m [19] | CSPNeXt-m | KR | 76.9 | 90.9 | 80.7 | 68.9 | 81.9 |
| RTMPose-l [19] | CSPNeXt-l | KR | 75.9 | 90.8 | 85.6 | 69.8 | 84.9 |
| HEGPosEs | ResNet-50 | KR | 79.0 | 90.9 | 80.9 | 68.9 | 84.8 |
| HEGPosEs | Swin-T | KR | 80.9 | 93.8 | 84.8 | 69.9 | 81.9 |
| HEGPosEs | Swin-L | KR | 84.6 | 93.9 | 85.9 | 70.8 | 85.8 |

**Table 4** Performance of HEGPosEs on MPII Human Pose in end-to-end methods

| Method | Backbone | Loss | mAP | mAP_50 | mAP_75 | mAP_Medium | mAP_Large |
|---|---|---|---|---|---|---|---|
| Mask R-CNN [40] | ResNet-50 | HM | 66.5 | 87.5 | 71.1 | 61.3 | 73.4 |
| Mask R-CNN [40] | ResNet-101 | HM | 66.6 | 87.4 | 74.0 | 61.5 | 74.4 |
| PRTR [16] | ResNet-50 | KR | 68.6 | 88.2 | 75.2 | 66.2 | 76.2 |
| HrHRNet [11] | HRNet-w32 | HM | 67.6 | 86.2 | 73.0 | 61.5 | 79.9 |
| InsPose [13] | ResNet-50 | KR + HM | 64.9 | 86.9 | 68.9 | 59.5 | 70.9 |
| HEGPosEs | ResNet-50 | KR + HM | 75.0 | 88.8 | 78.9 | 64.0 | 79.4 |
| HEGPosEs | ResNet-101 | KR + HM | 78.9 | 90.5 | 77.9 | 67.1 | 79.0 |
| HEGPosEs | Hourglass-104 | KR + HM | 80.8 | 88.6 | 70.9 | 58.8 | 76.5 |

**Table 5** Performance of HEGPosEs on MPII Human Pose in non-end-to-end methods

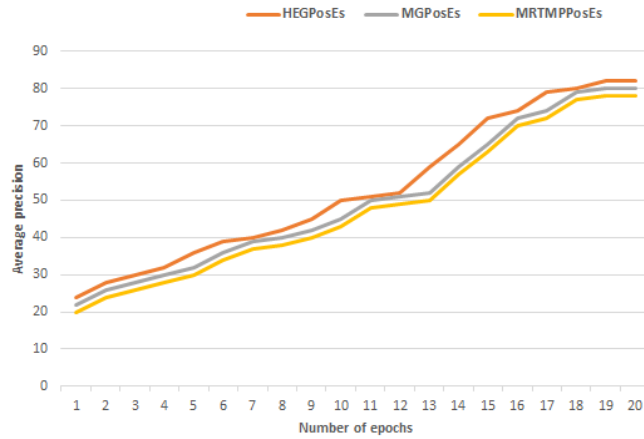

**Fig. 10** Convergence curve comparison of HEGPosEs, MGPosEs and MRTMPPosEs

| Method | With Human Detection Detector | 12e | 24e | 36e | 48e | 60e |
|---|---|---|---|---|---|---|
| ED-Pose [33] | Yes | 60.5 | 67.5 | 69.7 | 70.8 | 71.6 |
| GroupPose [18] | No | 61.0 | 67.6 | 70.1 | 71.4 | 72.0 |
| GroupPose [18] | Yes | 61.4 | 68.1 | 70.3 | 71.6 | 72.2 |
| RTMPose [19] | No | 60.7 | 67.4 | 70.0 | 71.3 | 71.7 |
| RTMPose [19] | Yes | 61.9 | 67.5 | 70.2 | 71.4 | 71.9 |
| HEGPosEs | No | 62.7 | 68.4 | 73.0 | 73.3 | 73.7 |
| HEGPosEs | Yes | 64.9 | 69.5 | 73.3 | 73.4 | 73.9 |

**Table 6** The model convergence analysis

| Method | Input Resolution | Frames Per Second ↑ | Time [milliseconds] ↓ |
|---|---|---|---|
| PETR [13] | 480 × 800 | 20.0 | 50 |
|  | 800 × 1333 | 12.1 | 83 |
| QueryPose [32] | 480 × 800 | 19.0 | 56 |
|  | 800 × 1333 | 13.4 | 75 |
| ED-Pose [33] | 480 × 800 | 42.4 | 24 |
|  | 800 × 1333 | 24.7 | 40 |
| GroupPose [18] | 480 × 800 | 68.6 | 15 |
|  | 800 × 1333 | 31.3 | 32 |
| RTMPose [19] | 480 × 800 | 68.5 | 14 |
|  | 800 × 1333 | 31.2 | 31 |
| HEGPosEs | 480 × 800 | 78.6 | 10 |
|  | 800 × 1333 | 40.3 | 31 |

**Table 7** The model inference speed analysis

## 5 Experiments and Analysis

In order to further strengthen our hypothesis we present additional experimental results. Considering baselines ensemble learning methods are compared. This helps us to understand performance specific impacts. The mean average precision (mAP) evaluation metric [20] is used to validate results. We have used 128-bit OS having x128 Intel processor with RAM of 32 GB. The method is implemented with Python 3.11.0 in Google Colab with 8 GPUs NVIDIA A100 hardware. With multiple performance scores on benchmark datasets stacking and bagging pipelines are being setup. We stress upon significant test conditions [20]. The hybrid ensemble combines prediction results from each individual component which reduces predictions based variance and generalization errors. The solution used here considers committee of methods which gives good fit for each component. With different hyper parameters all components have dissimilar configurations. The predictions are performed for each developed model. The actual predictions are achieved as average of predictions. The different ensemble combinations are used in order to reach best results [20]. During experimentation process we also varied major elements of ensemble methods [20]. These include ensemble models, their combinations and training data. The training data is varied with k-fold cross-validation and bagging with bootstrap aggregations. The models are varied with multiple runs of training alongwith hyper parameters tuning, snapshots, vertical representations and horizontal epochs. The combinations are varied with average and weighted average of models, generalization of stacks and other averages. We do not have single best ensemble method. The experiments are performed on publicly available human pose estimation benchmarked dataset viz. MPII Human Pose [10] which is state-of-the-art dataset and evaluates articulated human poses. It has 40000 images of which 25,000+ images are annotated with 420 human activities and movements. YouTube videos helped us to extract images. The annotated and unannotated frames are included here. There is rich annotation data, occlusions in body movements, 3D and torso orientation in test data. The model performance for MPII Human Pose dataset is shown in Table 3 and performance curves with confusion matrix in Fig. 9. In Table 4 performance of HEGPosEs on MPII Human Pose datasets for end-to-end methods is shown. In Table 5 performance of HEGPosEs on MPII Human Pose datasets for non-end-to-end methods is shown. For both end-to-end and non-end-to-end methods different backbones and loss functions are used. The backbones used include ResNet-50, ResNet-101, Swin-L, Swin-T, CSPNeXt-t, CSPNeXt-s, CSPNeXt-m, CSPNeXt-l, HRNet-w32 and Hourglass-104. The loss functions used include heatmap regression (HM), keypoint regression (KR), HM + KR, bone radius (BR) + residual log-likelihood regression (RLE) and BR + KR. With standard evaluation process, we have thresholds denoted as mAP, mAP_50, mAP_75, mAP_Medium and mAP_Large. The thresholds mAP_Medium and mAP_Large represent medium and large object sizes respectively. The results have confirm the fact that stated pose estimation method is more robust to occlusion with improvement in dense regression accuracy. Fig. 10 compares convergence curves of HEGPosEs, MGPosEs and MRTMPPosEs. The model convergence analysis is presented in Table 6 with 12, 24, 36, 48 and 60 epochs. The model inference speed is shown in Table 7.

## 6   Conclusion

In this research work hybrid ensemble group pose estimation method HEGPosEs is developed to estimate authentic human poses. Here robust pose estimation objective is reached with pose transformation PoseTrans. HEGPosEs is initialized with pre-trained models. Transfer learning with early stopping have been used for model efficiency and fast training. The pose transformation method results in best pose estimates in conjugation with pose estimator. The ensemble is being evaluated on test datasets with respect to various parameters. This method's viability is being established based on comparative analysis of group pose estimation methods conducted on benchmarked datasets. It makes pose estimation method more adaptive to occlusion and improves dense regression accuracy. The proposed pose estimation method can be further tested with additional datasets such as CrowdPose, 3DPW, LSPe, AMASS, VGG, Human3.6M etc which can be taken up as future work. The estimation of poses is significant for four important reasons. It identifies human body movements which impacts their overall health condition. The study of human body movements leads to human poses. It helps in detection and classification of body joints. The set of coordinates with respect to each body joint is captured. These are known as keypoints which describe posture of person. These connection forms a pair. With human pose estimation models we can dynamically track points in real-time motion. The results have confirmed application of this method in several real time pose estimation applications viz clinical sciences, human activity, media and sports with considerable improvements.